\documentclass[runningheads,a4paper]{llncs}

\usepackage{amssymb,amsmath,algorithm,algorithmic}
\usepackage{graphicx}
\newfloat{Algorithm}{thp}{}[section]
\newfloat{Algorithm}{tb}{lox}
\floatname{Algorithm}{Algorithm}
\usepackage{url}
\newcommand{\eat}[1]{}
\urldef{\mailsa}\path|{ssrajan@yahoo-inc.com,shirish@csa.iisc.ernet.in|
\newcommand{\keywords}[1]{\par\addvspace\baselineskip
\noindent\keywordname\enspace\ignorespaces#1}

\begin{document}
\def\Mu{{\mbox{\boldmath $\mu$}}}
\def\Beta{{\bf w}}
\def\bSigma{{\bf \Sigma}}
\def\beta{w}
\def\calD{{\cal D}}
\def\bX{{\bf X}}
\def\bx{{\bf x}}
\def\by{{\bf y}}
\def\bm{{\bf m}}
\def\bxo{{\bf x}_1}
\def\bxt{{\bf x}_2}
\def\bxi{{\bf x}_i}
\def\bxj{{\bf x}_j}
\def\bxn{{\bf x}_N}
\def\bbf{{\bf f}}
\def\bbfhat{\hat{\bf f}}
\def\bz{{\bf 0}}
\def\bbl{{\bf l}}
\def\bK{{\bf K}}
\def\bA{{\bf A}}
\def\bM{{\bf M}}
\def\bL{{\bf L}}
\def\bI{{\bf I}}
\def\bu{{\bf u}}
\def\buc{{\bf u}^c}
\def\bui{\bar{\bf u}_i}
\def\buj{\bar{\bf u}_j}
\def\bPi{{\mbox{\boldmath $\Pi$}}}
\def\bfalp{{\mbox{\boldmath $\alpha$}}}
\def\bfgamma{{\mbox{\boldmath $\gamma$}}}
\def\bfSigma{{\mbox{\boldmath $\Sigma$}}}
\def\bfmu{{\mbox{\boldmath $\mu$}}}
\def\bfnu{{\mbox{\boldmath $\nu$}}}
\def\bfbeta{{\mbox{\boldmath $\beta$}}}
\def\bftheta{{\mbox{\boldmath $\theta$}}}
\def\bfeta{{\mbox{\boldmath $\eta$}}}
\def\bfzeta{{\mbox{\boldmath $\zeta$}}}
\def\kuu{{{\bf K}_{{\bf u},{\bf u}}}}
\def\kuf{{{\bf K}_{{\bf u},{\bf f}}}}
\def\kfu{{{\bf K}_{{\bf f},{\bf u}}}}
\def\kui{{\bf K}_{{\bf u},i}}
\def\kiu{{\bf K}_{i,{\bf u}}}
\def\kuj{{\bf K}_{{\bf u},j}}
\def\kju{{\bf K}_{j,{\bf u}}}
\def\bfAu{{\bf A}_{\bf u}}
\def\muii{mu_{-i}(i)}
\def\bfu{{\bf u}}
\def\bfl{\bf L}
\def\bfui{{\bar \bf u}_{i}}
\def\bfuj{{\bar {\bf u}}_{j}}
\def\bful{{\bar {\bf u}}_{l}}
\def\bfAu{{\bf A}_{\bfu}}
\def\bfinvGu{{\bf G}^{-1}_{\bfu}}
\def\bfGu{{\bf G}_{\bfu}}
\def\bfinvGut{{\bf G}^{-T}_{\bfu}}
\def\bfGut{{\bf G}^T_{\bfu}}
\def\bfinvGuj{{\bf G}^{-1}_{\bfuj}}
\def\bfGujt{{\bf G}^T_{\bfuj}}
\def\bfGuj{{\bf G}_{\bfuj}}
\def\bfinvGujt{{\bf L}^{-T}_{\bfuj}}
\def\bfLujt{{\bf L}^T_{\bfuj}}
\def\bfinvLu{{\bf L}^{-1}_{\bfu}}
\def\bfLu{{\bf L}_{\bfu}}
\def\bfinvLut{{\bf L}^{-T}_{\bfu}}
\def\bfLut{{\bf L}^T_{\bfu}}
\def\bfinvLuj{{\bf L}^{-1}_{\bfuj}}
\def\bfLuj{{\bf L}_{\bfuj}}
\def\bfinvLujt{{\bf L}^{-T}_{\bfuj}}
\def\bfLujt{{\bf L}^T_{\bfuj}}
\def\bfAuj{{\bf A}_{\bfuj}}
\def\kuuj{{{\bf K}_{{\bfuj},{\bfuj}}}}
\def\bfzjt{{\bf z}^T_j}
\def\bfzlt{{\bf z}^T_l}
\def\bfhjt{{\bf h}^T_j}
\def\bfhj{{\bf h}_j}
\def\bfhi{{\bf h}_i}
\def\bfzj{{\bf z}_j}
\def\bfzl{{\bf z}_l}
\def\bfzi{{\bf z}_i}
\def\dj{d_j}
\def\di{d_i}
\def\ej{e_j}
\def\ei{e_i}
\def\kfl{{\bf K}_{{\bf f},l}}
\def\kfi{{\bf K}_{{\bf f},i}}
\def\kfj{{\bf K}_{{\bf f},j}}
\def\kfl{{\bf K}_{{\bf f},l}}
\def\kif{{\bf K}_{i,{\bf f}}}
\def\kjf{{\bf K}_{j,{\bf f}}}
\def\klf{{\bf K}_{l,{\bf f}}}
\def\bfbjt{{\bf b}^T_j}
\def\bfblt{{\bf b}^T_l}
\def\cj{c_j}
\def\cl{c_l}
\def\bfbj{{\bf b}_j}
\def\bfbl{{\bf b}_l}
\def\bfwjt{{\bf w}^T_j}
\def\bfwj{{\bf w}_j}
\def\bfwi{{\bf w}_i}
\def\kii{{\bf K}_{i,i}}
\def\kil{{\bf K}_{i,l}}
\def\kij{{\bf K}_{i,j}}
\def\kji{{\bf K}_{j,i}}
\def\kli{{\bf K}_{l,i}}
\def\kjj{{\bf K}_{j,j}}
\def\kll{{\bf K}_{l,l}}
\def\bfy{{\bf y}}
\def\uj{u_j}
\def\ui{u_i}
\def\etaj{\eta_j}
\def\etai{\eta_i}
\def\bfs{\bf S}
\def\bfsi{{\bf S}_{-i}}
\def\dmax{d_{\rm max}}
\def\nsv{{n_{\textsc{sv}}}}
\def\xtil{x}  
\def\ftil{\tilde{f}}
\def\L{{\cal{L}}}
\def\ker{k}   

\mainmatter  

\title{An Additive Model View to Sparse Gaussian Process Classifier Design}

\titlerunning{An Additive Model View to Sparse Gaussian Process Classifier Design}

%
%
\author{
Sundararajan S\inst{1} \and Shirish Shevade\inst{2}}
\institute{Yahoo! Labs, Bangalore, India,\\ \email{ssrajan@yahoo-inc.com}
\and Department of Computer Science and Automation \\ Indian Institute of Science\\ Bangalore, India\\ \email{shirish@csa.iisc.ernet.in}}
%
\authorrunning{Sundararajan and Shirish}


%
%

\toctitle{Lecture Notes in Computer Science}
\tocauthor{Authors' Instructions}
\maketitle

\begin{abstract}
We consider the problem of designing a sparse Gaussian process classifier (SGPC) that generalizes well. 
Viewing SGPC design as constructing an additive model like in boosting, we present an efficient and 
effective SGPC design method to perform a stage-wise optimization of a predictive loss function. 
We introduce new methods for two key components viz., site parameter estimation and basis vector selection 
in any SGPC design. The proposed adaptive sampling based basis vector selection method aids in 
achieving improved generalization performance at a reduced computational cost. This method can also be used 
in conjunction with any other site parameter estimation methods. It has similar computational and storage 
complexities as the well-known information vector machine and is suitable for large datasets. The hyperparameters 
can be determined by optimizing a predictive loss function. The experimental results show better generalization 
performance of the proposed basis vector selection method on several benchmark datasets, particularly for relatively
smaller basis vector set sizes or on difficult datasets.
\end{abstract}

\newpage
\section{Introduction}
\label{introduction}
Sparse Gaussian Process (GP) classifier design aims at addressing the issues of high computational and storage costs associated with learning a full model GP ($O(n^3)$ and $O(n^2)$ respectively)\cite{Rasmussen06} using $n$ training examples, and involves using a representative data set, called the {\em basis vector} set, from the input space. In this way, the computational and memory requirements are reduced to $O(n d^2_{max})$ and $O(n d_{max})$ respectively, where $d_{max}$ is the size of the basis vector set ($d_{max} \ll n$). Further, the costs of predictive mean and variance computations for an example are reduced from $O(n)$ and $O(n^2)$ to $O(d_{max})$ and $O(d^2_{max})$ respectively. 

In this work, we focus on developing an efficient Sparse Gaussian Process Classifier (SGPC) design algorithm. Several approaches have been proposed in the literature to design sparse GP classifiers. These include on-line GP learning \cite{Csato02} and entropy or information gain based Informative Vector Machine (IVM) \cite{Lawrence03,Seeger07}. Particularly relevant to this work is IVM which is inspired by the technique of assumed density filtering (ADF)~\cite{Minka01,Csato02}. In general, an SGPC design algorithm using the ADF approximation involves site parameter estimation, basis vector selection and hyperparameter optimization. While the site parameters are estimated using a moment matching technique in the ADF approximation, hyperparameters are estimated by optimizing marginal likelihood or negative logarithm of predictive probability (NLP)~\cite{Rasmussen06}. Different methods to select the basis vectors include entropy, information gain and validation based methods~\cite{Shirish08}. Experimental comparisons of the IVM with entropy based method and validation based method on various benchmark datasets showed that though the IVM method is efficient, it does not generalize well particularly on difficult datasets, and it requires more number of basis vectors to achieve similar generalization performance compared to the validation based method. Though the validation based method generalizes well, it is computationally expensive. Therefore, there is a need to have an efficient algorithm to design SGPCs that generalize well. 

{\bf Contributions:} 
Viewing SGPC design as construction of an additive model (that is, a linear combination of basis functions)~\cite{Friedman00}, a basis vector addition can be seen as adding a basis function in each iteration like in boosting~\cite{Schapire98}. With this view we introduce new methods to select the basis vectors and, estimate their site parameters by optimizing a predictive loss function. These estimated site parameters determine the coefficient of the basis function in the additive model. Further, an adaptive sampling based basis vector selection method is proposed, which aids in effective basis vector selection and computational cost reduction. The proposed basis vector selection method has same computational complexity as used by IVM. We also compare the generalization performance of various basis vector selection methods. Experimental results show that the proposed method gives comparable or better performance on a wide range of real-world large datasets. In particular, the proposed method is significantly better compared to the entropy and information gain based methods for relatively smaller $d_{max}$ values or on difficult datasets.

The paper is organized as follows. Section 2 presents an SGPC design algorithm with the ADF approximation. The proposed methods and implementation aspects are given in Section 3. Section 4 covers related work. Experimental results are presented in Section 5 and the paper concludes with Section 6.

\section{GP and Sparse GP Classification}
Given a training data set with input-output pairs $\calD=\{{\bf x}_i, y_i\}^{n}_{i=1}$ where ${\bf x}_i$ $\in$ $R^d$ and $y_i$ $\in$ $\{+1,-1\}$, 
the goal is to design a GP classifier that generalizes well. In standard GPs for classification \cite{Rasmussen06}, true function value at each ${\bf x}_i$ is represented as a latent random variable $f({\bf x}_i)$. Let us denote $f({\bf x}_i)$ by $f_i$. The prior distribution of $\{{\bf f}({\bf X}_n)\}$ is a zero mean multivariate joint Gaussian, denoted as ${\it p}({\bf f})\:=\:{\mathcal N}(\cdot;{\bf 0},{\bf K})$, where ${\bf f}\:=\:[f_1,\ldots,f_n]^T$, ${\bf X}_n\:=\:[{\bf x}_1,\ldots,{\bf x}_n]$, and ${\bf K}$ is an $n \times n$ covariance matrix whose $(i,j)^{th}$ element is $k({\bf x}_i,{\bf x}_j)$.
An example covariance function is the squared exponential function: $k({\bf x}_i,{\bf x}_j)\:=\:v_0\:\exp(-\frac{1}{2} \sum_{m=1}^d \frac{{(x_{i,m}-x_{j,m})^2}}{\sigma^2})$. Here, $v_0$ and $\sigma^2$ denote the signal variance and kernel width respectively. 
In this work we use the probit noise model, $p(y_i|f_i,\lambda,b)$ = $\Phi(\lambda y_i(f_i+b))$ where $\Phi(\cdot)$ is the cumulative distribution of the standard Gaussian ${\mathcal N}(\cdot;0,1)$ with zero mean and unit variance, the slope of which is controlled by $\lambda$($>$0) and $b$ is a bias hyperparameter. With independent, identical distribution assumption, we have ${\it p}({\bf y}|{\bf f},\bfgamma)\:=\:\prod_{i=1}^n p(y_i|f_i;\bfgamma)$ where $\bfgamma = [\lambda,\:\;b]$. Let $\bftheta \:=\:[v_0, \sigma^2, \bfgamma]$ denote the hyperparameters that characterize the GP model. With these modeling assumptions, the expressions for latent posterior and predictive distributions are available~\cite{Rasmussen06}. 
In SGPC design using the ADF approximation \cite{Lawrence03}, a factorized form of $q_{\bu}(\bf f|\calD,\bftheta)$ (given below) is made use of, to build an
approximation to $p({\bf f}|{\calD},\bftheta)$ in an incremental fashion.
Let $\bu$ denote the index set of the training examples which are included 
in the approximation. Then we have 
\begin{equation}
q_{\bu}({\bbf}|\calD,\bftheta) \propto {\mathcal N}(\bbf;{\bf 0},{\bK}) \prod_{i \in {\bu}} \exp\left\{
-\frac{p_i}{2} {(f_i - m_i)}^2 \right\}
\label{qi}
\end{equation}
and $p({\bf f}|{\calD},\bftheta) \approx q_{\bu}({\bf f}|{\calD},\bftheta) = {\cal N}(\bbf;\hat{\bf f}, {\bf A})$ where ${\bf A} = {({\bf K}^{-1}+ \bPi)}^{-1}$ and $\hat{\bf f} = {\bf A} \bPi {\bm}$, ${\bf m} = {(m_1,\ldots,m_n)}^T$ and $\bPi = {\rm diag}{(p_1,\ldots,p_n)}$. The parameters $m_i$ and $p_i$, $i=1\rightarrow n$ are called the site function parameters and the set ${\bu}$ is called the {\em active} or {\em basis vector} set. Note that $\bu$ is actually associated with the inputs $\bX_{\bu}$. We refer to $\bu^c=\{1,2,\ldots,n\} \setminus {\bu}$ as the non-active set. In practice, the active set size $|\bu|$ is restricted by the user specified parameter, $d_{max}$. Note that the site function parameters corresponding to $\bu^c$ are zero. Thus a SGPC model is defined by the basis vector set ${\bu}$, its associated site function parameters $({\bf m}_{\bu},\bPi_{\bu})$ and the hyperparameters $\bftheta$. In general, SGPC design algorithms differ with respect to the basis vector selection, site parameter estimation and hyperparameters optimization methods. A typical SGPC design algorithm using the ADF approximation is given in Algorithm 1. 

\begin{algorithm}
   \caption{SGPC Design}
   \begin{algorithmic}
 \STATE 1. Initialize the hyperparameters $\bftheta$. Set $d_{max}$, $tol$, $iter_{max}$ and, $iter$=0.
 \REPEAT
 \STATE 2. Initialize ${\bf A} := {\bf K}, \bu = \{\}, \buc = \{ 1,2,\ldots,n\}, \hat{f}_i = p_i = m_i = 0 \; \forall \; i \in \buc$. $iter=iter+1$.
 \REPEAT
 \STATE 3. Select a basis vector $j$ from $\buc$ as per the chosen basis vector selection method.
 \STATE 4. Update the site parameters $p_j$, $m_j$, posterior mean ($\hat{\bf f}$) and
variance ($\mbox{diag}(\bA)$). 
 \STATE 5. Set $\bu = \bu \cup \{j\}$ and   $\buc = \buc \setminus \{j\}$. 
 \UNTIL{$|\bu| = d_{max}$}
 \STATE 6. Re-estimate the hyperparameters $\bftheta$ by optimizing a suitable loss function, keeping $\bu$ and the corresponding site parameters constant.
 \UNTIL{$iter = iter_{max}$ or change in the loss function value $<~tol$}
 \vspace{-0.05in}
 \end{algorithmic}
\end{algorithm}
We now briefly describe the ADF approximation method \cite{Lawrence03} to implement step 4. Suppose that an example index $j$ is added to the current basis vector set ${\bu}$. Let ${\bar \bu}_j=\bu \cup \{j\}$. After updating the site function parameters $p_j$ and $m_j$, incremental calculations are carried out to update $\bbfhat$ and $\mbox{diag}(\bA)$ corresponding to ${\bar \bu}_j$. This is achieved by maintaining two matrices $\bL$ and $\bM$ where $\bL$ is the lower-triangular Cholesky factor of ${\bf B}={\bI} + \bPi^{1/2}_{\bu,\bu} {\bK}_{\bu,\bu} \bPi^{1/2}_{\bu,\bu}$ and
${\bM} = {\bL}^{-1} \bPi_{\bu,\bu} {\bK}_{\bu,\cdot}$\footnote{\tiny{The subscript, $(\bu,\bu)$, of a matrix is used to represent the rows and columns of the matrix corresponding to the elements of the set $\bu$. The subscript, $(\bu,.)$ denotes the rows of the matrix corresponding to the elements of the set $\bu$.}}. Note that $\bA = \bK - {\bM}^T {\bM}$. However, only the diagonal elements of $\bA$ are needed in the algorithm and are updated as given in (\ref{eq:eqn5}) below. Assuming $\lambda=1$, with $z_j  =  \frac{y_j (\hat{f_j}+b)}{\sqrt{1+ A_{jj}}}, \; 
\alpha_j = \frac{y_j {\mathcal N}(z_j;0,1)}{\Phi(z_j)} \sqrt{\frac{1}{1+A_{jj}}}, \;
 \nu_j = \alpha_j \left( \alpha_j + \frac{(\hat{f}_j+b)}{1+A_{jj}} \right)$ the site function parameters are updated as: 
\begin{equation}
p_j = \frac{\nu_j}{1-A_{jj} \nu_j},\;\; m_j = \hat{f_j} + \frac{\alpha_j}{\nu_j}.
\label{eq:eqn2}
\end{equation}
Let ${\bbl}  = \sqrt{p_j} {\bM}_{\cdot,j},\;\; l = \sqrt{1+p_j \bK_{j,j} - {\bbl}^T{\bbl}}, \;\; \Mu = {l}^{-1} (\sqrt{p_j} \bK_{\cdot,j} - {\bM}^T {\bbl})$. Then $\bM$ is updated by appending the row vector ${\Mu}^T$ and $\bL$ is updated by appending $[{\bL}\:{\bf 0}]$ with $[{\bbl}^T\:l]$. 
The posterior variance and mean are updated as:
\begin{equation}
\mbox{diag}(\bA) := \mbox{diag}(\bA) - \Mu^2, \;\; \hat{\bbf} := \hat{\bbf} + \alpha_j l  p^{-1/2}_j \Mu.
\label{eq:eqn5}
\end{equation}
In (\ref{eq:eqn5}), ${\Mu}^2$ denotes squaring of each element in $\Mu$.
In the outer loop the hyperparameters are optimized by maximizing the marginal likelihood (ML)~\cite{Lawrence03}, $q_{\bu}({\bf y}|{\bf X},\bftheta)=\int p({\bf y}|{\bf f},\bfgamma) q_{\bu}({\bf f}|\calD,\bftheta) d{\bf f}$ or minimizing the negative logarithm of predictive probability (NLP) loss (under cumulative Gaussian noise model)~\cite{Shirish08},
\begin{equation}
\mbox{NLP}({\bu}, \bftheta) = -\frac{1}{|{\buc}|} \sum_{i \in \buc} \log \Phi\left(\frac{y_i (\hat{f}_i+b)}{\sqrt{1+{\bA}_{ii}}}\right).
\label{nlp}
\end{equation}
Finally the predictive target distribution for an unseen input $x_*$ is given by: 
$q_{\bu}(y_\ast|{\bx}_\ast) = \Phi\left( \frac{y_\ast (\hat{f}_\ast + b)}{\sqrt{1+\sigma^2_\ast}}\right)$
 where ${\hat f}_*={\bf k}_{*,\bu}\bPi^{1\over2}_{\bu}{\bf B}^{-1}\bPi^{1\over 2}_{\bu}{\bf m}_{\bu}$ and $\sigma^2_{*}=k({\bf x}_*,{\bf x}_*)-{\bf k}_{*,\bu}\bPi^{1\over2}_{\bu}{\bf B}^{-1}\bPi^{1\over 2}_{\bu}{\bf k}_{\bu,*}$. In the next section we propose new methods for effective basis vector selection and site parameters optimization (steps 3 and 4 in Algorithm 1). 

\section{Proposed Methods}
Friedman et al~\cite{Friedman00} showed how boosting~\cite{Schapire98} can be seen as a way of fitting an additive model, 
		$f_{M}({\bf x}) = \sum_{m=1}^M \beta_m \:\psi({\bf x};\delta_m)$
where $\beta_m$, $m=1,2,\ldots,M$ are the expansion coefficients, and $\psi({\bf x};\delta_m) \in {\mathcal R}$ are the basis functions characterized by the parameters $\delta_m$, $m=1,2,\ldots,M$, and $M$ is the number of basis functions ($M=d_{max}$). This model is fit by minimizing a loss function averaged over the training data, that is:
$\min_{\{\beta_m,\delta_m\}^M_{1}} \sum_{i=1}^n exp(-y_i f_M({\bf x}_i))$
where an exponential loss function is used. In forward stagewise additive modeling the basis functions are added one at a time and, the coefficient and the basis function parameter $(\beta_m,\delta_m)$ are optimized by keeping the coefficients and parameters of the previously chosen basis functions constant. That is, $\{\beta_m,\delta_m\}= arg \min_{\beta,\delta} exp(-y_i f_{m-1}({\bf x}_i)+\beta \psi({\bf x};\delta)), m>1$. Friedman et al~\cite{Friedman00} also presented a related loss function that is based on the binomial likelihood, given by: $\sum_{i=1}^n \log(1+exp(-2y_i f_M({\bf x}_i)))$. 

With this view, we consider the SGPC design as constructing a forward stagewise additive model. 
Before we show the equivalences between the selection of a basis function and its coefficient to the selection of a basis vector $(j)$ and its site parameters ($p_j,m_j$), we define an objective function (called predictive loss function) that we propose to use to select a basis function and its coefficient in each iteration of the SGPC design algorithm:  
\begin{equation}
\mbox{NLP}_{a}(\{\bu \cup j\}, \bftheta) = - {1 \over n} \sum_{i=1}^{n} \log \Phi\left(\frac{y_i (\hat{f}_i+b)}{\sqrt{1+{\bA}_{ii}}}\right) 
\label{nlpboost1}
\end{equation}
where $j \in \buc$ and, ${\hat f}_i, {\bA}_{ii}$ are computed using $\{\bu \cup j\}$. This objective function has a behavior similar to the exponential and binomial likelihood loss functions mentioned above and, is also an upper bound on the training set error. That is, we have:
\begin{equation}
	{1 \over n} |\{i: sgn({\hat f}({\bf x}_i)+b) \ne y_i\}| \;\le\; {1 \over {\log(2) }} \;\mbox{NLP}_{a}({\bu}, \bftheta) 
\label{inequality}
\end{equation}
Here the left hand side represents the training set error. The inequality follows from noting that $0\le \Phi(z)\le 1$, $\Phi(0)=0.5$ and that $-\log(\Phi(z))$ monotonically decreases in the interval $(-\infty,\infty)$. Note that $\log(\Phi(0))=-\log(2)$ and is required for appropriate scaling so that $-{{\log(\Phi(z))} \over {\log(2)}}\ge 1\; when \;z\le 0$. Thus (\ref{nlpboost1}) is an upper bound on the training set error. 

\begin{figure}
\begin{center}
{
  \vskip -0.2in
	\includegraphics[width=6cm,height=4cm]{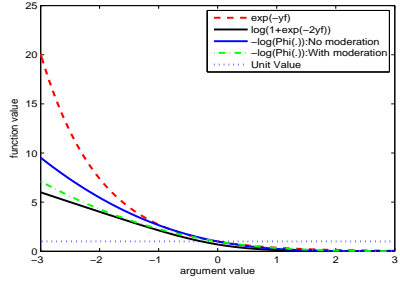}
}
\vskip -0.05in
\caption{{\scriptsize Exponential, binomial log likelihood, $-{{\log(\Phi(\cdot))} \over{\log(2)}}$ functions with and without moderation. In the moderation case, the variance was set to 0.5. Zero variance corresponds to no moderation. A reference function that takes unit value is also shown.}}
\vspace{-0.4in}
\label{FigA}
\end{center}
\end{figure}

\noindent{{\bf Comparison of Objective Functions:}} Firstly, unlike the exponential loss function used in boosting, the function $\Phi(\cdot)$ is not separable. That is, the linear combination of basis functions that appear inside $\Phi(\cdot)$ cannot be written as a product of individual terms. This separability property of the exponential function is useful for the interpretation of building successive weak classifiers on the training data with weighted distribution. However, keeping all the previous basis functions with the associated coefficients fixed and, optimizing over {\it only} an additional basis function along with its coefficient using (\ref{nlpboost1}) essentially has the same desirable effect. It may be noted that like (\ref{nlpboost1}), the binomial log likelihood is not separable in strict sense (without any approximation). Secondly, the GP classifier has the advantage of providing predictive variance information which is useful in moderating the predictive probability. Specifically when the uncertainty or variance is large, this probability gets reduced accordingly. This is very important particularly when the data points are sparse in a certain region of the input space or when the data is noisy. Thus, use of (\ref{nlpboost1}) would be more robust. The behaviors of $-\log(\Phi(\cdot))$ with and without moderation along with the other loss functions are shown in Figure~\ref{FigA}. 

\noindent{\bf Forward Stagewise Additive Model View:} We now show using (\ref{eq:eqn5}) that the SGPC design using Algorithm 1 with (\ref{nlpboost1}) as the objective function (to select the basis vectors and their coefficients) is equivalent to building a forward stagewise additive model. In particular, a basis vector selection results in a basis function choice and the coefficient optimization essentially results in its site parameters estimation in each iteration (steps 3 and 4 in Algorithm 1). Note that the notions of stage and iteration in Algorithm 1 are equivalent. First, let us look at the steps 3 and 4 of Algorithm 1 more closely. After selecting a basis vector $j$ and updating its site parameters $(p_j,m_j)$ at the $t$-th iteration, the following posterior variance and mean update can be obtained by simplifying (\ref{eq:eqn5}): 
\begin{equation}
\mbox{diag}(\bA)^{(t+1)} := \mbox{diag}(\bA)^{(t)} - \eta_j \tilde{\bf k}^2_{.,j}, \;\; \hat{\bbf}^{(t+1)} := \hat{\bbf}^{(t)} + {\tilde \alpha}_j \tilde{\bf k}_{.,j}
\label{eq:eqn6}
\end{equation}
where $\tilde{\bf k}_{.,j}=({\bf k}_{.,j}-{\bf k}_{.,\bu_{t}}\bPi^{1\over2}_{\bu_{t}}{\bf B}^{-1}_{\bu_{t}}\bPi^{1\over2}{\bf k}_{\bu_{t},.})$ and $\bu_t$ is the basis vector set at the $t$-th iteration. Here $\eta_j={{p_j}\over{1+p_j{\bA}^{(t)}_{jj}}}$ and ${\tilde{\alpha}}_j=\eta_j (m_j-{\hat f}^{(t)}_j)$. Note that $\eta_j\ge 0$. Then the process of adding the $j$th basis vector is equivalent to adding a basis function $\tilde{\bf k}({\bf x},{\bf x}_j)$. That is, we can define the additive model function for SGPC as: ${\hat f}^{(t+1)}({\bf x}) = {\hat f}^{(t)}({\bf x}) + {\tilde \alpha}_j {\tilde k}({\bf x},{\bf x}_j)$. Here, ${\tilde k}({\bf x},{\bf x}_j)=k({\bf x},{\bf x}_j)-{\bf k}({\bf x},{\bf x}_{\bu_{t}})\bPi^{1\over2}_{\bu_{t}}{\bf B}^{-1}_{\bu_{t}}\bPi^{1\over2}{\bf k}({\bf x}_{\bu_{t}},{\bf x})$ (where ${\bf k}({\bf x},{\bf x}_{\bu_{t}})$ is a row vector of size $|\bu_{t}|$), and is dependent on the input ${\bf x}_j$ through $k({\bf x},{\bf x}_j)$, the previously chosen functions and their site parameters. Note that in both the ADF approximation and the proposed methods, the site parameters of the previously selected basis vectors are not updated whenever a new basis vector is added. This is done to reduce the computational complexity. Next, we can see that the choice of $\tilde {\alpha}_j$ is dependent on the site parameters $m_j$ and $p_j$. This is because ${\hat f}^{(t)}_j$ and ${\bA}^{(t)}_{jj}$ are fixed once the $j$th basis vector is chosen. Now, relating ${\hat f}^{(t+1)}({\bf x})$ to the predictive mean vector in (\ref{eq:eqn6}), we see that the predictive mean vector is nothing but the evaluation of the function ${\hat f}^{(t+1)}({\bf x})$ for the training inputs ${\bf x}_i, i=1,\ldots,n$. Therefore, selection of the $j$th basis vector and estimation of its site parameters ($p_j,m_j)$ in each iteration (stage) of the SGPC design algorithm  essentially determine the basis function ${\tilde k}({\bf x},{\bf x}_j)$ and its coefficient ${\tilde \alpha}_j$. To summarize, we have the final classifier function (excluding the bias hyperparameter $b$) and the predictive variance on an input ${\bf x}$ as: 
\begin{eqnarray}
{\hat f}({\bf x})=\sum_{i=1}^{d_{max}} {\tilde \alpha}_i {\tilde k}({\bf x},{\bf x}_i) \\
{\hat \sigma}^2({\bf x}) = k({\bf x},{\bf x})-\sum_{i=1}^{d_{max}} \eta_i {\tilde k}({\bf x},{\bf x}_i)
\label{eq:sgpc}
\end{eqnarray}
Note that the expression for ${\hat \sigma}^2({\bf x})$ follows from the expression for $\mbox{diag}(\bA)^{(t+1)}$ on the left hand side of (\ref{eq:eqn6}). It is interesting to see that the variance is a non-increasing function as more and more basis functions are added. Having shown the equivalence, we next show how the $j$th basis function and the associated coefficient $\tilde {\alpha}_j$ can be obtained by optimizing (\ref{nlpboost1}) in each iteration. As we have seen before, the choice of a basis vector determines the basis function and we describe next how this selection is done. 

\noindent{\bf Basis Vector Selection Method:} From efficiency viewpoint, we propose to select a basis vector as:
\begin{equation}
	j =\arg\min_{i \in {\bf J}} \mbox{NLP}_{a}(\{\bu \cup i\}, \bftheta).
\label{iboost}
\end{equation}
where {\bf J}, a working set, is a randomly chosen subset of $\buc$, $|{\bf J}|$=min($\kappa$,$|\buc|$) and $\kappa$ can be set to 59~\cite{Smola01}. To select one basis vector using (\ref{iboost}) the computational cost is $O(\kappa nd_{max})$. Therefore a method to reduce the factor $\kappa$ without significantly degrading generalization performance will be very useful. We achieve this by changing the sampling strategy (from random sampling) used to construct the working set ${\bf J}$. In the proposed adaptive sampling technique, we construct ${\bf J}$ by sampling from $\buc$ according to a distribution that changes after a basis vector is added in each iteration. The sampling distribution is given by:
\begin{equation}
	\chi^{(t+1)}_{j\in \buc_t} = {1 \over {V^{(t)}}} \Bigl(1 - \Phi\bigl({{y_j ({\hat f}^{(t)}_j+b)} \over \sqrt{{1+{\bf A}^{(t)}_{jj}}}}\bigr)\Bigr) 
\label{distribution}
\end{equation}
where $V^{(t)}$ is a normalizing constant. Here, ${\hat {\bf f}}^{(t)}_j$ and ${\bf A}^{(t)}_{jj}$ are computed using the basis vectors in $\bu_t$. Since ${\hat f}$ and ${\bf A}$ change after inclusion of every basis vector in the inner loop, the distribution also changes and the sampling becomes adaptive.  

To understand why such a sampling along with (\ref{iboost}) would be useful, we can see that if $\Phi(\cdot) \rightarrow 1$ (for a correctly classified example with high predictive probability), then the probability of selecting such an example as a basis vector will be relatively small. On the other hand, the probability of selecting a misclassified example with low predictive probability (that is, $\Phi(\cdot) \rightarrow 0$) will be relatively high. We found that selecting the most violated example (that is, the example with the least $\Phi(\cdot)$ in $\buc$) in each iteration results in poor basis vector selection for noisy and difficult datasets. The adaptive sampling technique can safeguard against such a selection and is robust across different datasets. Next, the sign of ${\tilde \alpha}_j$ in (\ref{eq:eqn6}) gets adjusted in such a way that ${\hat {\bf f}}^{(t+1)}$ moves in the desired direction for a given $\tilde {\bf k}_{.,j}$. This desired movement is expected to happen for all the examples having same class label that are close enough to the $j$th example. Therefore, with a choice of an example (having low value of $\Phi(\cdot)$), ${\hat {\bf f}}^{(t+1)}$ moving in the desired direction and variance $diag({\bf A})^{(t+1)}$ non-increasing, we expect the NLP value in (\ref{nlpboost1}) to improve particularly for the examples with wrong predictions or low predictive probability. In this sense the basis vector selection using (\ref{iboost}) and (\ref{distribution}) tends to mimic the selection of a base classifier in boosting~\cite{Schapire98} that minimizes the training set error with weighted distribution. This helps in getting a better generalization performance for a fixed $\kappa$ compared to random sampling. Alternatively, $\kappa$ can be reduced to get the same generalization performance. Experimental results support these claims.       

\noindent{\bf Site Parameters Optimization Method:} Having constructed the working set {\bf J} using the adaptive sampling technique,  we optimize (\ref{nlpboost1}) to find $\tilde{\alpha}_i$ for each basis vector $i\in {\bf J}$. As shown earlier optimizing over $\tilde {\alpha}_i$ is equivalent to optimizing over the site parameters $m_i$ and $p_i$ for a given basis vector. Essentially we have a two dimensional ($m_i,p_i)$ non-linear optimization problem. Note that it is a constrained optimization problem (under certain condition given below) since the posterior variance $diag(\bA)^{(t+1)}$ should be non-negative after every iteration. Assuming that $diag(\bA)^{(t)}$ is non-negative it turns out that $p_i$ must satisfy: $ {\tilde {\eta}}_i \ge \eta_i={{p_i}\over{1+p_i{\bf A}^{(t)}_{ii}}}$ where $\tilde {\eta}_i=min_l\;\{{{\bA^{(t)}_{ll}} \over {{\tilde k}^2_{l,i}}}\}$. On further simplification we find that if $\tilde {\eta}_i{\bA}^{(t)}_{ii}\ge 1$ then we have an unconstrained optimization problem (in $\tau_i$ when we work with $p_i=\exp(\tau_i)$); otherwise we have a constrained optimization problem with $0\le p_i \le {{\tilde {\eta}_i} \over {1-\tilde {\eta}_i{\bA}^{(t)}_{ii}}}$. This can be solved using any standard nonlinear optimization technique. {\it To summarize, we construct ${\bf J}$ using adaptive sampling, optimize $\tilde {\alpha}_i,\;\forall i\in {\bf J}$ and select the basis vector using (\ref{iboost}).}

\begin{figure*}
\begin{center}
{
	\includegraphics[width=5.5cm,height=5cm]{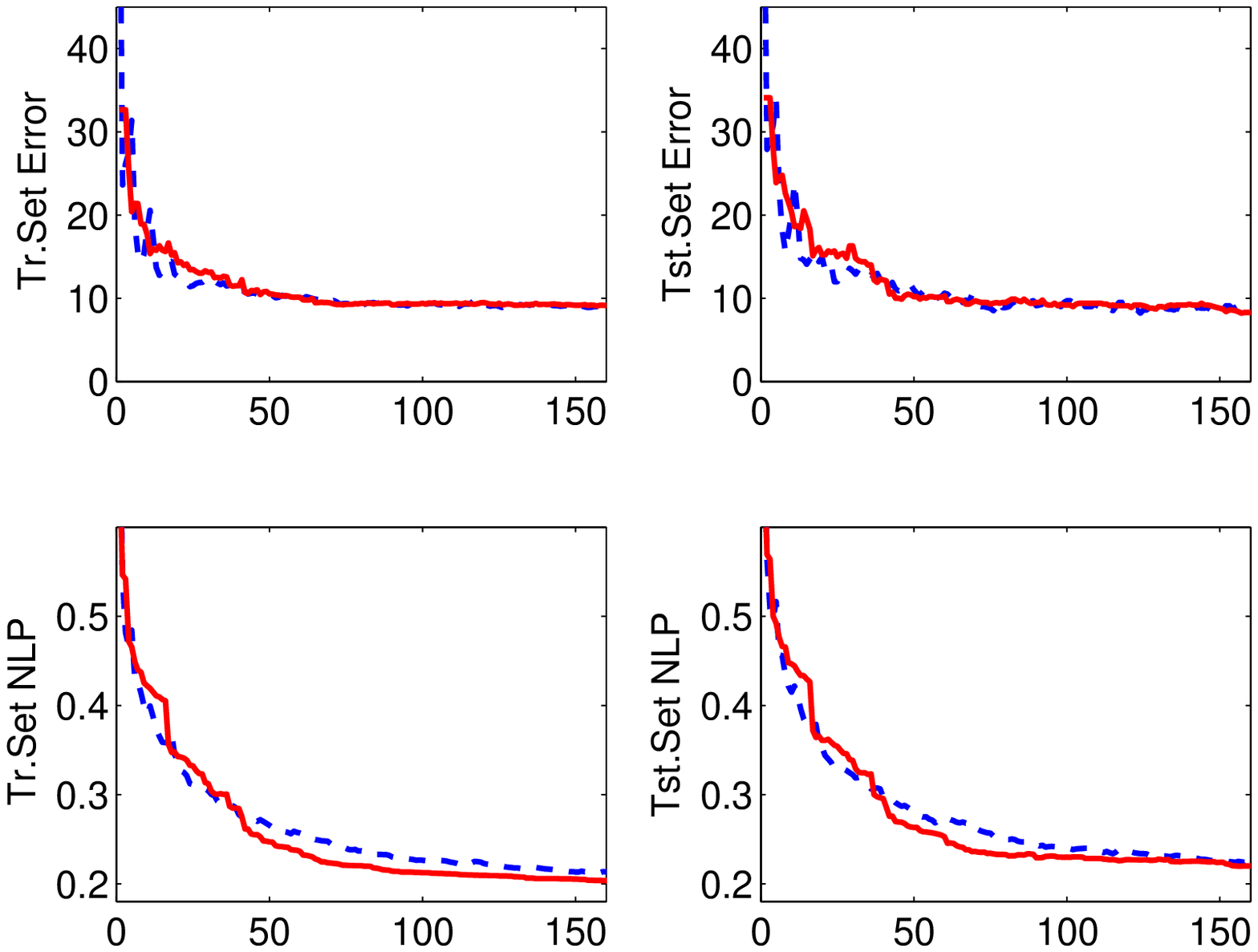}
}
\hskip 0.1in
{
	\includegraphics[width=5.5cm,height=5cm]{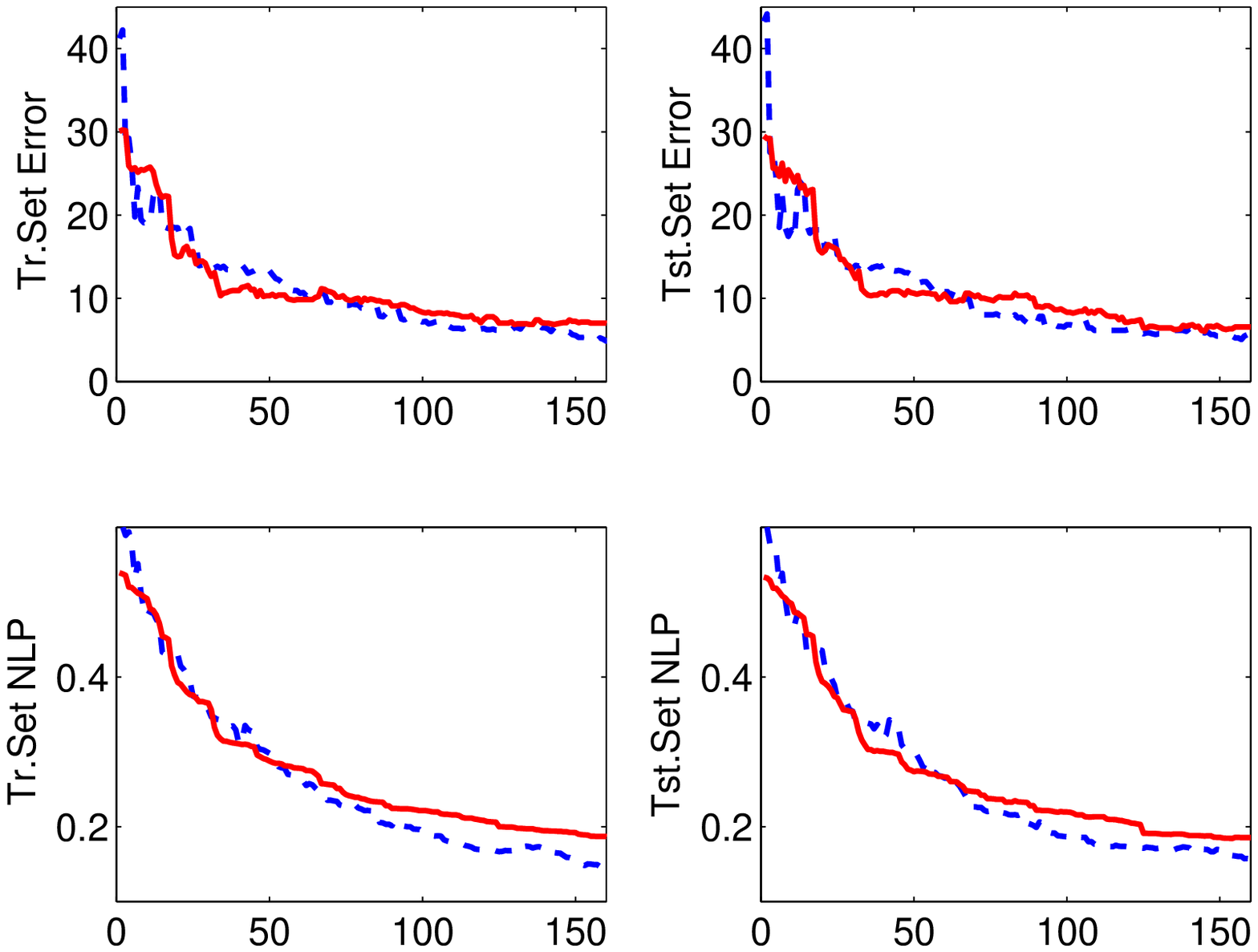}
}
\vskip -0.15in
\caption{{\scriptsize The left panel (group of 4 plots) corresponds to Waveform dataset and the right panel corresponds to Image dataset. The first and second rows show the training/test set errors and NLP loss as the basis vectors are added in the inner loop just before termination. The solid-red and dashed-blue lines correspond to $\tilde{\alpha}$ with moment matching and constrained optimization cases with ADF approximation. In this experiment we set $\kappa=2$.}}
\vspace{-0.4in}
\label{Fig2}
\end{center}
\end{figure*}
\section{Related work}
In this section we briefly describe three closely related methods that we compare with the proposed method.
In entropy based method~\cite{Lawrence03}, a basis vector is chosen according to the change in the entropy of the posterior process (\ref{qi}) after inclusion in the model and is given by: $j=\arg\min_{i\in\buc} \log({\bar \lambda}_i)$ where ${\bar \lambda}_i=1-\nu_i{\bf A}_{ii}$. In information gain based method~\cite{Seeger07}, a basis vector is chosen according to the information gain (which is defined as negative of the Kullback-Leibler divergence) obtained from the posterior process after inclusion in the model and is given by: $j=\arg\min_{i\in\buc} \{-\log({\bar \lambda}_i)+{1\over{{\bar \lambda}_i}}+{{({\hat f}^{'}_{i}-{\hat f}_i)^2} \over {\bf A}_{ii}}\}$. Here ${\hat f}_i$ and ${\hat f}^{'}_i$ denote the predictive mean before and after the inclusion of the $i$th basis vector. Compared to the entropy based selection, this method takes the predictive mean also into account and differs from the way ${\bar \lambda}_i$ is traded-off between the first and second term. Both these methods are very efficient since the relevant quantities that are needed to compute the appropriate measure for the basis vector selection are maintained throughout in the inner loop of Algorithm 1. Both these methods maximize the marginal likelihood for hyperparameter optimization. 

In validation based method~\cite{Shirish08}, a working set  ${\bf J} \subseteq \buc$ of fixed size $\kappa$ ($\kappa=\min({|\buc|}, 59))$ is constructed by sampling randomly from $\buc$. The basis vector that minimizes (\ref{nlp}) is chosen and this involves computation of a new NLP value after inclusion for each $i\in {\bf J}$. Thus the computational cost for one basis vector selection is $O(\kappa n d_{max})$. The hyperparameters are selected by minimizing (\ref{nlp}). 

While all the three methods use moment matching with the ADF approximation to estimate the site parameters (\ref{eq:eqn2}), the proposed site parameters optimization method provides an alternate way to estimate these parameters. Note that one can also use (\ref{eq:eqn2}) in conjunction with the proposed adaptive sampling based basis vector selection method. On comparing the objective functions used by the proposed method and validation based method, we see that the form of (\ref{nlp}) is same that of (\ref{nlpboost1}) except that the summation happens only over $\buc$. While the validation based method viewed (\ref{nlp}) as obtaining the NLP performance estimate with a validation set, (\ref{nlpboost1}) is motivated from the additive modeling viewpoint and, minimizing an upper bound on the training set error. Furthermore, the validation based method uses fixed uniform sampling instead of adaptive sampling. Note that the difference between (\ref{nlp}) and (\ref{nlpboost1}) is expected to be insignificant when $d_{max}\ll n$ (usually the case in SGPC design with large datasets) and this condition is important to avoid any overfitting.  
\section{Experiments}
The summary of the datasets used in the experiments is given in Table 1. These datasets are part of Gunnar Raetsch's benchmark datasets available at \url{http://theoval.cmp.uea.ac.uk/~gcc/matlab/default.html
}. We changed the training and test set sizes of the top five datasets in Table 1 to demonstrate the effectiveness of the proposed method on large datasets. For the first four datasets we picked top $3600$ test examples from the original test set partition and added to the training set. The remaining examples were used as the test set. Note that this construction however results in reduction of the test set size. In the case of {\it Splice} dataset, we picked the top $1000$ examples from the test set partition. The modified train and test set sizes are shown Table 1. We considered only the first 25 partitions of the first four datasets. In all the experiments we used the squared exponential covariance function and Algorithm 1 described in Section 2. A conjugate gradient method was used to optimize (\ref{nlp}) (unless otherwise specified) in the outer loop for optimizing the hyperparameters, and $iter_{max}$ was set to 20. We kept track of the best model based on the NLP loss value after every outer loop iteration. For comparison, we evaluated the test set error and NLP loss performance.          
\begin{table}
\begin{center}
\vspace{-0.25in}
\caption{\scriptsize{Datasets Description. $n$ and $m$ denote the training and test set sizes. $d$ and $pt$ denote the input dimension and number of partitions.}}
\begin{small}
\begin{tabular}{|l|c|c|c|c|} \hline 
 Dataset & $n$ & $m$ & $d$ & $pt$ \\ \hline
 \textsf{Banana} & 4000 & 1300 & 2 & 25 \\ \hline 
 \textsf{Waveform} & 4000 & 1000 & 21 & 25 \\ \hline
 \textsf{Twonorm} & 4000 & 3400 & 20 & 25 \\ \hline
 \textsf{Ringnorm} & 4000 & 3400 & 20 & 25 \\ \hline
 \textsf{Splice} & 2000 & 1175 & 60 & 20 \\ \hline
 \textsf{Image} & 1300 & 1010 & 18 & 20 \\ \hline
\end{tabular}
\vspace{-0.4in}
\end{small}
\end{center}
\label{Datasets}
\end{table}

We conducted three experiments. Due to the space constraints we present only selected results. In the first experiment we illustrate the effectiveness of the proposed method of site parameters (equivalently, $\tilde{\bfalp}$) optimization. The results on one partition of the \textsf{Waveform} and \textsf{Image} datasets are shown in Figure 2. This method is compared against using (\ref{eq:eqn2}) for site parameters optimization. Although some minor variations were seen between the two methods, statistical analysis showed that the performance differences were not significant. Thus, the constrained optimization is an effective alternate method to estimate the site parameters. We now discuss certain practical aspects of this optimization. During optimization, the variance can become zero (within numerical accuracy), for some choice of the hyperparameter values and, also due to the greedy nature of the basis vector selection method. While this can be handled in some way (for example by exiting the inner loop), optimizing over individual $\tilde {\alpha}_i$'s can become slightly expensive for large datasets. Note that the function and gradient computations are linear in $n$. We can control the optimization cost by restricting the number of function and gradient evaluations with some inaccuracy in the solution. Therefore, the proposed optimization is also efficient. 

\begin{figure}
\vskip -0.1in
\begin{center}
{
	\includegraphics[width=5.5cm,height=5.5cm]{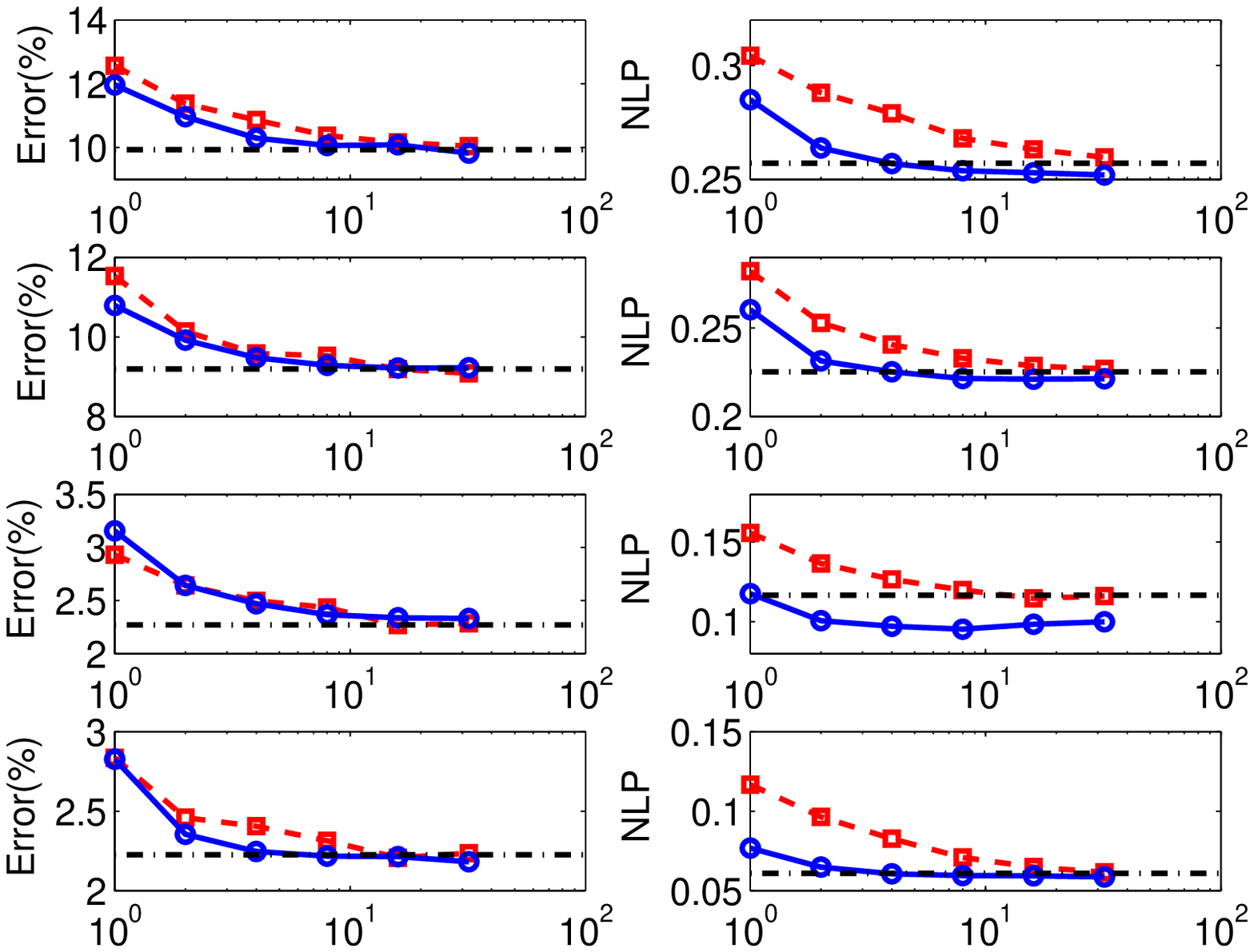} 
}
\hskip 0.1in
{
	\includegraphics[width=5.5cm,height=5.25cm]{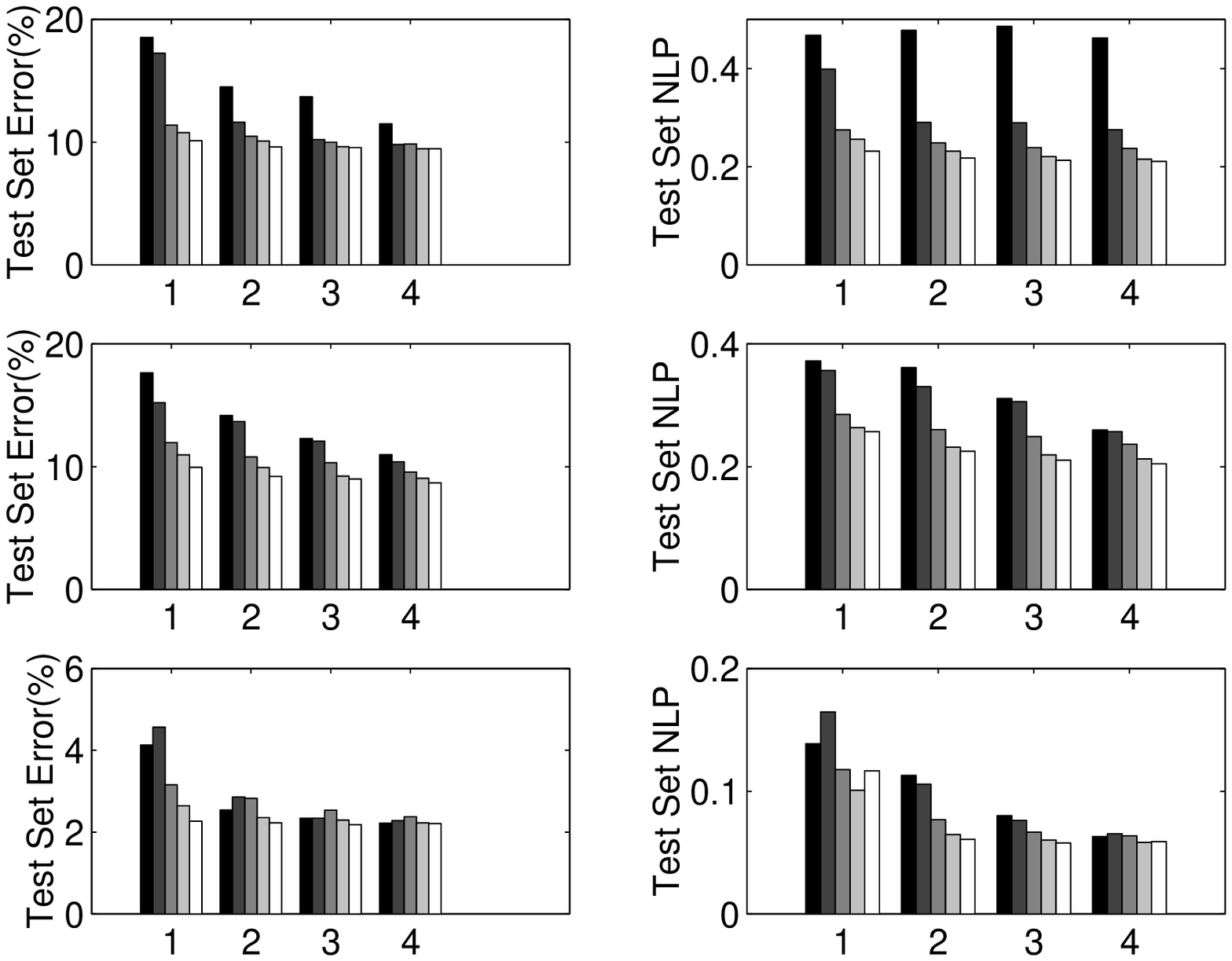}
}
\vskip -0.1in
\caption{{\scriptsize \textsf{\bf {Left Panel of eight plots}}: Test set error and NLP loss performance of the random sampling (dashed-red-square) and adaptive sampling (solid-blue-circle) methods for different values of $\kappa$. The dashed-dot-black line corresponds to the validation based method with $\kappa$=59. Top two rows correspond to \textsf{Waveform} dataset for $d_{max}$=40 and 80 (in that order). The bottom rows correspond to \textsf{Twonorm} dataset for $d_{max}$=40 and 80. \textsf{{\bf Right Panel of six plots}}: Test set performance of the various basis vector selection methods (entropy, information-gain, proposed method with $\kappa$=1 and 2, and validation based method ($\kappa$=59) (different gray shades) in that order) for different values of $d_{max}$ (40, 80, 160 and 320 correspond to the x-axis values of 1, 2, 3 and 4 respectively). Each row corresponds to one dataset. The results on \textsf{Banana}, \textsf{Waveform} and \textsf{Twonorm} datasets are given in that order.}}
\label{fig3}
\end{center}
\end{figure}

In the next two experiments, we kept the site parameter estimation (using (\ref{eq:eqn2})) and the hyperparameter estimation (using (\ref{nlp})) same, and only changed the basis vector selection method in the step 3 of Algorithm 1. This is because our goal here is to compare the quality of the different basis vector selection methods. First, we demonstrate the effectiveness of the adaptive sampling method in the basis vector selection. This is done by comparing it with random (uniform) sampling method. We conducted this experiment on all the datasets given in Table 1. The test set error and NLP loss performance results on two datasets are given in Figure~\ref{fig3} (left panel) for two different values of $d_{max}$. These results were obtained by averaging the performance over the partitions. We found that the adaptive sampling method consistently performed better across all the datasets, particularly with respect to the NLP loss measure. This is because the choice of the basis vectors made by the adaptive sampling method is based on the predictive distribution. We also observed improved test set error performance on several cases. It was also observed that the performance difference reduces as the working set size $\kappa$ increases. It can also be seen that $\kappa$ value of 2 is sufficient for the adaptive sampling method to get similar NLP generalization performance as the validation based method (see the second column in the left panel of Figure~\ref{fig3}). 

In the third experiment, we compared the performance of the proposed method, validation based method, entropy and information gain based basis vector selection methods. In the case of proposed method, we evaluated the performance with $\kappa=$ 1 and 2, thus ensuring that the complexity for the basis vector selection is the same as that of the entropy and information gain based methods. We conducted this experiment for four different values of $d_{max}$ (40, 80, 160 and 320) on all the datasets given in Table 1. The test set error and NLP loss performance on three datasets are shown in Figure~\ref{fig3} (right panel). They were obtained by averaging the performance over the partitions. We compared the performance of various methods using statistical significance tests. We first conducted Wilcoxon test on the test set error and NLP loss obtained from the partitions, on each dataset. All the observations from the tests below are made at the significance level of 0.05. The results indicated better test set error and NLP performance of the proposed method over the entropy and information gain based methods on almost all the datasets. Specifically we observed that the proposed method performed better on difficult datasets (relatively higher test set errors) like \textsf{Banana}, \textsf{Waveform} and \textsf{Splice} for all values of $d_{max}$ with respect to (w.r.t.) both the measures. On \textsf{Twonorm} and \textsf{Ringnorm} datasets it performed better w.r.t. the NLP loss measure for all the values of $d_{max}$. While it performed better than the entropy based method on the \textsf{Ringnorm} dataset for all values of $d_{max}$ w.r.t. the test set error, the performance was the same at higher values of $d_{max}$ in other cases. The information gain based method performed better than the entropy based method on the \textsf{Banana}, \textsf{Waveform} and \textsf{Ringnorm} datasets. The entropy based method performed better than the information gain based method w.r.t. the test set error in the case of \textsf{Twonorm} dataset. We observed that the entropy based method performed better than all the methods at lower values of $d_{max}$ on the \textsf{Image} dataset. On comparing the proposed method ($\kappa=$ 1 and 2) with the validation based method, we found that the validation based method performed better w.r.t. the test set error at lower values of $d_{max}$ (40 and 80). 

Next, following \cite{Demsar06}, we conducted Friedman's test with six datasets (Table 1) and four methods, namely, the proposed method (with $\kappa$=2), validation, entropy and information gain based methods. To conduct this test, we used the average test set error and NLP values obtained from averaging over the partitions. The p-values obtained for the test set error and NLP measure were (0.02, 0.04, 0.04, 0.39) and (0.002, 0.002, 0.01, 0.09) respectively for four different values of $d_{max}$ (40, 80, 160 and 320) in that order. When $d_{max}$ was 320, the results were not significantly different w.r.t. both the measures. Since the null hypothesis was rejected for $d_{max}$ values of 40, 80 and 160, we next conducted the Bonferroni-Dunn post-hoc test to compare the proposed method with the other three methods. This test revealed that there were no significant differences between the proposed and validation based methods for all values of $d_{max}$ w.r.t. both the measures. On comparing the proposed method with the entropy and information gain based methods, we found that while the results were not significantly different w.r.t. the test set error, they were significant w.r.t. the NLP measure for lower $d_{max}$ values at 0.1 level. Overall, it was seen that the p-value became larger and the performance differences across the methods reduced as $d_{max}$ was increased. 

Except for the validation based method ($\kappa=$ 59), all the methods required almost the same computational time for the basis vector selection. An approximate timing measurement of one inner loop (for $d_{max}$=80) showed that the proposed method with $\kappa=$ 1 took approximately 20 seconds for the \textsf{Banana} dataset (on a machine with 2 GB of RAM and dual core Intel CPU running at $1.83$ GHz). In general, we found that the proposed method was $5$ times faster than the validation based method on almost all the datasets. This comparison was based on the Matlab implementations of these methods. The speed improvement was not as high as $59$. We believe that efficient matrix based operations in Matlab helped the validation based method significantly and, expect the speed improvement to be higher with implementations in other programming languages like C.  

\section{Conclusion}
We considered the problem of designing an SGPC from an additive model estimator viewpoint. We introduced new methods for basis vector selection and site parameters estimation based on the predictive loss function. An adaptive sampling method that aids in effective basis vector selection and computational complexity reduction was proposed. The proposed basis vector selection method has same computational and storage complexities as that used by IVM and, is thus suitable for large datasets. The experimental results showed better generalization performance of the proposed method on several benchmark datasets, particularly for relatively smaller $d_{max}$ values or on difficult datasets. 

\bibliography{esgpc}
\bibliographystyle{abbrv}

\end{document}